\pdfoutput=1

\documentclass[11pt]{article}

\usepackage[]{acl}

\usepackage{times}
\usepackage{latexsym}

\usepackage[T1]{fontenc}

\usepackage[utf8]{inputenc}

\usepackage{microtype}

\usepackage{graphicx}
\usepackage{amsmath}
\usepackage{amssymb}
\usepackage{amsfonts}
\usepackage{inconsolata}
\usepackage{multirow}
\usepackage{marvosym}
\usepackage{longtable}
\usepackage{xcolor}
\usepackage{booktabs}

\newcommand\askip{\hspace{10pt}}

\title{Zero-shot Triplet Extraction by Template Infilling}

\author{Bosung Kim\textsuperscript{\Neptune}\thanks{~~The work was partially done when Bosung Kim was a research intern at Megagon Labs.}\askip Hayate Iso\textsuperscript{\Cancer}\askip Nikita Bhutani\textsuperscript{\Cancer}\\
\textbf{Estevam Hruschka\textsuperscript{\Cancer}}\askip %
\textbf{Ndapa Nakashole\textsuperscript{\Neptune}} \askip \textbf{Tom Mitchell}\textsuperscript{\Cancer\Leo} \\
  \textsuperscript{\Neptune}University of California, San Diego\askip 
  \textsuperscript{\Cancer}Megagon Labs \askip
  \textsuperscript{\Leo}Carnegie Mellon University\askip \\
  {\tt bosungkim@ucsd.edu}\askip  \texttt{nnakashole@eng.ucsd.edu}\\
  \texttt{\{hayate,nikita,estevam,tom\}@megagon.ai}
  }

\begin{document}
\maketitle
\begin{abstract}

{The task of triplet extraction aims to extract pairs of entities and their corresponding relations from unstructured text. Most existing methods train an extraction model on training data involving specific target relations, and are incapable of extracting new relations that were not observed at training time. Generalizing the model to unseen relations typically requires fine-tuning on synthetic training data which is often noisy and unreliable. We show that by reducing triplet extraction to a template infilling task over a pre-trained language model (LM), we can equip the extraction model with zero-shot learning capabilities and eliminate the need for additional training data. We propose a novel framework,  \textsc{ZETT} (\textbf{ZE}ro-shot \textbf{T}riplet extraction by \textbf{T}emplate infilling), that aligns the task objective to the pre-training objective of generative transformers to generalize to unseen relations. Experiments on FewRel and Wiki-ZSL datasets demonstrate that ZETT shows consistent and stable performance, outperforming previous state-of-the-art methods, even when using automatically generated templates.\footnote{The code is available at \url{https://github.com/megagonlabs/zett}}}%

\end{abstract}

\section{Introduction}

\begin{figure}
    \centering
     \includegraphics[width=1.0\linewidth]{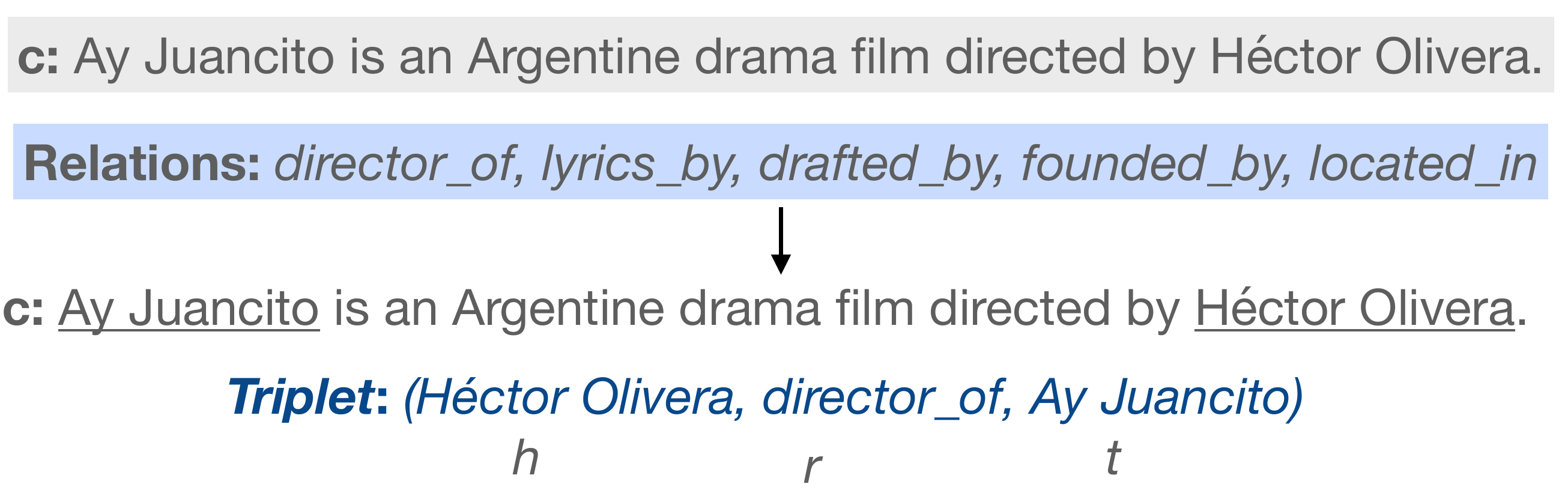}
    \caption{Triplet extraction task: given a context $c$ and a set of relations $\mathcal{R}$, extract pairs of entities and their relations. Zero-shot extraction aims at extracting relations not covered in the training data.}
    \label{fig:fig1}
\end{figure}

Extracting pairs of entities and their relations from unstructured text is vital to several applications including knowledge base population, text retrieval and question answering \cite{10.5555/2886521.2886624,xu-etal-2016-question}. Traditional approaches obtain entity pairs and relations step-by-step by considering entity recognition and relation classification as two separate sub-tasks. However, such multi-step approaches suffer from cascading errors and ignore interdependence between the tasks. Recent studies aim at extracting entities and relations together in a single step \citep{li-ji-2014-incremental,zheng-etal-2017-joint,tanl}. Given a set of pre-defined relations and an input text, they extract triplets of the form (head, relation, tail). We refer to this task as \textit{triplet extraction} (illustrated in Figure \ref{fig:fig1}).

If the relations are pre-defined, an extraction model can be trained on large-scale labeled data acquired via distant supervision or crowdsourcing \cite{sorokin-gurevych-2017-context,han-etal-2018-fewrel}. However, such methods are hard to adopt in real-world scenarios where ground-truth entities and relations cannot be specified in advance. To overcome these limitations, there is an increasing interest in generalizing models to extract entities and relations that are not observed during training --- a zero-shot setting.

Automatically generating training data for unseen relations is a widely used approach to render zero-shot capabilities to an extraction model. Distant supervision \citep{mintz-etal-2009-distant,zeng-etal-2015-distant,Ji_Liu_He_Zhao_2017} and data augmentation \citep{chia-etal-2022-relationprompt} can provide automatically labeled training data, but they suffer from quality and consistency of the synthetic data. They also require the model to be further fine-tuned on the synthetic data which can be computationally intensive. 
Yet another set of approaches extract unseen relations using cross-task knowledge learned from a collection of similar datasets and tasks \citep{zhong-etal-2021-adapting-language, sanh2022multitask} . However, the performance of these methods depends on the similarity between the datasets or the tasks.

Recent progress in large language models (LMs) such as GPT-3 shows that they can be adapted to zero-shot settings if the task objective is aligned with the pre-training objective \cite{NEURIPS2020_1457c0d6}. Based on this idea, many NLP tasks such as text classification \citep{zhong-etal-2021-adapting-language} and relation classification \citep{levy-etal-2017-zero, obamuyide-vlachos-2018-zero} have been successfully reformulated into prompt-based tasks.
{However, the state-of-the-art zero-shot triplet extraction approach still relies on synthetic data to generalize to unseen relations \citep{chia-etal-2022-relationprompt}. Moreover, most existing prompting approaches are designed for simple classification or generation tasks, making them unsuitable for the structured prediction such as triplet extraction which requires the identification of the complex triplet format. This work is the first study to explore how to reformulate triplet extraction into a prompt-based method to effectively and efficiently generalize to unseen relations.
}

We formulate triplet extraction as a template infilling task and propose a novel framework, \textsc{ZETT} (\textbf{ZE}ro-shot \textbf{T}riplet extraction by \textbf{T}emplate infilling) based on an end-to-end generative language model, T5~\citep{DBLP:journals/corr/abs-1910-10683}. Concretely, \textsc{ZETT} extends the input text with a relation template (e.g., "\texttt{<X>} is nominated for \texttt{<Y>}") and learns to generate the correct entity pair (e.g., "\texttt{<X>} John Bright \texttt{<Y>} Best Story") for the relation. In this manner, it aligns the task objective with the pre-training objective of the T5 model. The model is fine-tuned using annotated examples and a relation template for each relation in a set of predefined relations. Then at inference, the model extends the input text with template for each unseen relation and generates an entity pair and its score. It uses these scores to rank the relations and entity pairs and output the most-likely triplet(s). 

\begin{figure*}
    \centering
    \includegraphics[width=1.0\linewidth]{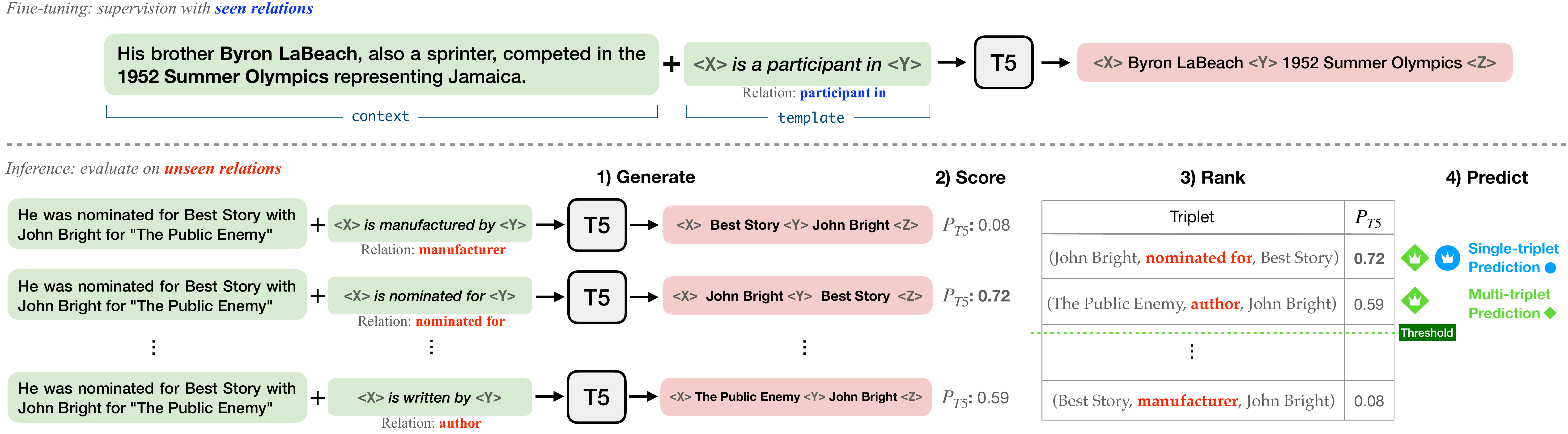}
    \caption{Fine-tuning and inference in ZETT: it fine-tunes T5 by extending input text with a relation template and learning to predict entity spans masked in the template. At inference, given templates for all unseen relation, it generates their entity spans and scores them. The best scoring sequences over a given threshold are then produced as the final output.}
    \label{fig:overview}
\end{figure*}

Although the model relies on a template for each seen and unseen relation, we show that \textsc{ZETT} can perform well even with automatically-generated templates for the relations. We further propose optimizations based on relation descriptions to improve efficiency at inference. Figure \ref{fig:overview} shows the overview of \textsc{ZETT}. Note that 
ZETT adopts an efficient single-step approach for the task that does not require synthetic data or additional fine-tuning for unseen relations. 

Experiments on publicly available datasets FewRel \citep{han-etal-2018-fewrel} and Wiki-ZSL \citep{chen-li-2021-zs} demonstrate that ZETT effectively generalizes to the zero-shot setting, outperforming state-of-the-art methods by up to 6 points in accuracy. We find that ZETT shows lower variance in performance compared to existing methods that rely on fine-tuning on noisy synthetic data. We also show that it is robust to the choice of template and can be integrated with automatically generated templates without significant loss in performance. In conclusion, ZETT is an effective and efficient method for the zero-shot triple extraction task.

\section{Method}

{
We now introduce \textsc{ZETT} (\textbf{ZE}ro-shot \textbf{T}riplet extraction by \textbf{T}emplate infilling), a generative triplet extraction framework that reformulates triplet extraction as template infilling.
}

\subsection{Task Definition}
{
The goal of triplet extraction is to extract  triplets $T = \{(h, r, t)\}$ given an input context $c$, where $h$ and $t$ are head and tail entities and $r$ is a predefined relation $r \in \mathcal{R}$.
In the zero-shot setting, we only have access to the dataset with the subset of relations $\mathcal{R}_{\text{seen}} \subset \mathcal{R}$ during training and have to generalize it to the dataset with unseen relations $\mathcal{R}_{\text{unseen}} \subset \mathcal{R}$ which is disjoint from seen relations: $\mathcal{R}_{\text{seen}}\cap \mathcal{R}_{\text{unseen}} = \phi$~\cite{chia-etal-2022-relationprompt}.
}

\subsection{Triplet Extraction by Template Infilling}
\label{sec:ZETT}

\begin{table*}[ht]
    \centering
    \small
    \resizebox{\linewidth}{!}{
    \begin{tabular}{p{0.75in}p{3in}p{2in}}
    \toprule
         \multicolumn{1}{c}{\textbf{Relation}} & \multicolumn{1}{c}{\textbf{Description}} & \multicolumn{1}{c}{\textbf{Template}} \\ \midrule
         participant in & event in which a person or organization was/is a participant & \texttt{<head>} is a participant in \texttt{<tail>}. \\ 
         publisher & organization or person responsible for publishing books, periodicals, printed music, podcasts, games or software & \texttt{<head>} is published by \texttt{<tail>}. \\ 
         screenwriter & person(s) who wrote the script for subject item & \texttt{<tail>} wrote the script for \texttt{<head>}. \\ 
        cast member & actor in the subject production& \texttt{<tail>} is an actor in \texttt{<head>}. \\
         \bottomrule
    \end{tabular}}
    \caption{Examples of relation types and templates. \texttt{<head>} and \texttt{<tail>} denote head and tail entities in a triplet. Each template is created based on its description in Wikidata \citep{10.1145/2629489}. We provide the full list of templates in supplementary materials.}
    \label{tab:templates}
\end{table*}

{
We reformulate the triplet extraction task to a template infilling task as follows.
For each relation $r \in \mathcal{R}$, we create a template $\tau_r$ with placeholders for the head (\texttt{<head>}) and tail (\texttt{<tail>}) entities, and then fill in these placeholders with entities from the context $c$ to identify the head ($h$) and tail ($t$) entities for the relation $r$.
For example, for the relation \textit{participant in}, we prepare a template, ``\texttt{<head>} is a participant in \texttt{<tail>}''. Using a context $c$ (``His brother Byron LaBeach, also a sprinter, competed in the 1952 Summer Olympics representing Jamaica.''), we fill the placeholders in the template to identify the head (``Byron LaBeach'') and tail (``1952 Summer Olympics'') entities for the relation \textit{participant in}.
We show more template examples in Table~\ref{tab:templates}.

{With this formulation, we can even extract unseen relations $\mathcal{R}_{\text{unseen}}$ simply by preparing their templates and infilling them. This eliminates the need for additional fine-tuning for the unseen relations. At inference, we perform template infilling for all target unseen relations and re-rank them based on the consistency between the context $c$ and the infilled template.}
}

\subsection{ZETT implementation with T5}

We instantiate ZETT using the pre-trained language model T5~\cite{DBLP:journals/corr/abs-1910-10683}.
T5 is pre-trained to predict consecutive spans randomly dropped out from a sentence, which is closely aligned with the template infilling task.

{As illustrated in Figure~\ref{fig:overview}, we build input-output pairs for the text infilling task using the context $c$, the gold triplet $T = (h, r, t)$, and corresponding template $\tau_r$.
We replace the placeholder tokens \texttt{<head>} and \texttt{<tail>} in the template $\tau_r$ with mask tokens \texttt{<X>} and \texttt{<Y>},\footnote{In T5 implementation, mask tokens are denoted as \texttt{<extra\_id\_n>}, where $n \in \{0, ..., 99\}$. We use simplified forms \texttt{<X>}, \texttt{<Y>}, and \texttt{<Z>} instead of \texttt{<extra\_id\_0>}, \texttt{<extra\_id\_1>}, and \texttt{<extra\_id\_2>}. We also note that \texttt{<X>} and \texttt{<Y>} are not respectively corresponding to \texttt{<head>} and \texttt{<tail>}. \texttt{<X>} and \texttt{<Y>} are used as mask tokens in the input sequence and as delimiters for predicted spans in the output sequence. Thus \texttt{<X>} always comes first followed by \texttt{<Y>}. On the other hand, the order of \texttt{<head>} and \texttt{<tail>} depends on its template.}
and concatenate it with the context $c$ to form a model input $x$. 
{Then, we fine-tune the model to learn to generate output $y$ consisting of the gold head and tail entities where each entity follows the corresponding mask tokens. We use the standard negative log-loss minimization $\mathcal{L} = -\log P_{\text{T5}}(y \mid x)$ for fine-tuning.}

}

\subsection{Inference with relation constraint}
\label{sec:relation_constraint}
At inference, we evaluate the model on the test data wherein input contexts have unseen relations $\mathcal{R}_{\text{unseen}}$ during training.
Given a context $c$, {we build multiple model inputs $\{x_r\}_{r \in \mathcal{R}_{\text{unseen}}}$}
by concatenating templates $\tau_{r}$ for each 
$r \in \mathcal{R}_\text{unseen}$.
We then generate entity tokens for each sequence. Since some contexts may have multiple triplets, we use beam search to generate multiple output sequences for each model input $x_r$.
We then compute a score for each output sequence as $P_{\text{T5}}(y \mid x_r)$. This score is used to rank all the generated triplets. The triplets are evaluated under single-triplet and multi-triplet settings. Under single-triplet setting, it is assumed that the input sentence has one triplet. In this setting, we predict the best scoring triplet as the output. Under multi-triplet setting, a sentence can have more than one triplet. In that case, we use threshold over the score to filter triplets. We tune the threshold on the validation set.

Exhaustively generating sequences for every unseen relation and scoring them can be inefficient in real-world scenarios where the number of unseen relations is large. Intuitively, not all unseen relations are related to the input context. 
Moreover, since ZETT relies on the LM's probability $P_{\text{T5}}$, the model tends to assign a higher score when the relation is frequently used in
common sentences even when it is not relevant to the context.
Therefore, instead of exhaustive scoring, we exploit relation constraints to filter out relations which are irrelevant to the given context.
Since we don't have any data for unseen relations, we adopt the relation extraction model that utilizes relation descriptions \citep{chen-li-2021-zs}. We use a sentence similarity score between the context $c$ and the description about the relation $r$ to exclude irrelevant relations from $\mathcal{R}_\text{unseen}$. For the relation descriptions, we use the descriptions from Wikidata as shown in Table \ref{tab:templates}.
Before generating entities, we first obtain the sentence embedding of the context and the relation's description using the off-the-shelf SBERT \citep{reimers-gurevych-2019-sentence}.
Then, we compute the cosine similarities between the context and relation's description embeddings and set the threshold $\delta$ on the validation set to filter out the relations whose similarity score is lower than $\delta$. After filtering out the irrelevant relations, we evaluate the model on the constrained unseen relation set. %

\section{Experiments}

\subsection{Dataset}

\begin{table}[t]
    \centering
    \small
    \resizebox{\linewidth}{!}{
    \begin{tabular}{cccc}
    \toprule
          & \# of examples & \# of relations & \# of entities   \\ \midrule
         FewRel & 56,000 & 80 & 72,954\\
         Wiki-ZSL & 94,383 & 113 & 77,623\\ \bottomrule
         & & & \\
    \end{tabular}}
    \begin{tabular}{ccccc}
    \toprule
          \multicolumn{2}{c}{$|\mathcal{R}_{\text{train}}|$}  & $|\mathcal{R}_{\text{test}}|$ & $|\mathcal{R}_{\text{validation}}|$ \\
          FewRel  & Wiki-ZSL & \multicolumn{2}{c}{$m$} & \\ \midrule
         70  & 103 & 5 & 5  \\
         65  & 98 & 10 & 5  \\
         60  & 93 & 15 & 5  \\ \bottomrule
    \end{tabular}
    \caption{Statistics of the datasets. $|\mathcal{R}|$ denotes the number of relations in each set, and $m$ is the number of relations in the test set.}
    \label{tab:stat_data}
\end{table}

We evaluate our method on two datasets: FewRel \citep{han-etal-2018-fewrel} and Wiki-ZSL \citep{chen-li-2021-zs}. FewRel is designed primarily for few-shot relation extraction. Wiki-ZSL is a subset of Wiki-KB and targets zero-shot relation extraction. Both datasets are created using distant supervision, but FewRel has  additionally been filtered by humans.
We use dataset versions released by \citet{chia-etal-2022-relationprompt} which have been transformed for zero-shot triplet extraction.
We follow their zero-shot set up for training and evaluation as follows:
1) we keep relation types in training, validation, and test splits disjoint, 2) we evaluate different methods under different settings for the size of unseen relation types ($m \in \{5, 10, 15\}$), 3). To avoid experimental noise, we repeat experiments using different data folds wherein relation types are split with different random seeds: \{0, 1, 2, 3, 4\}. 
Table \ref{tab:stat_data} shows statistics of each dataset and setting.

 \begin{figure*}[t]
    \centering
    \includegraphics[width=0.9\linewidth]{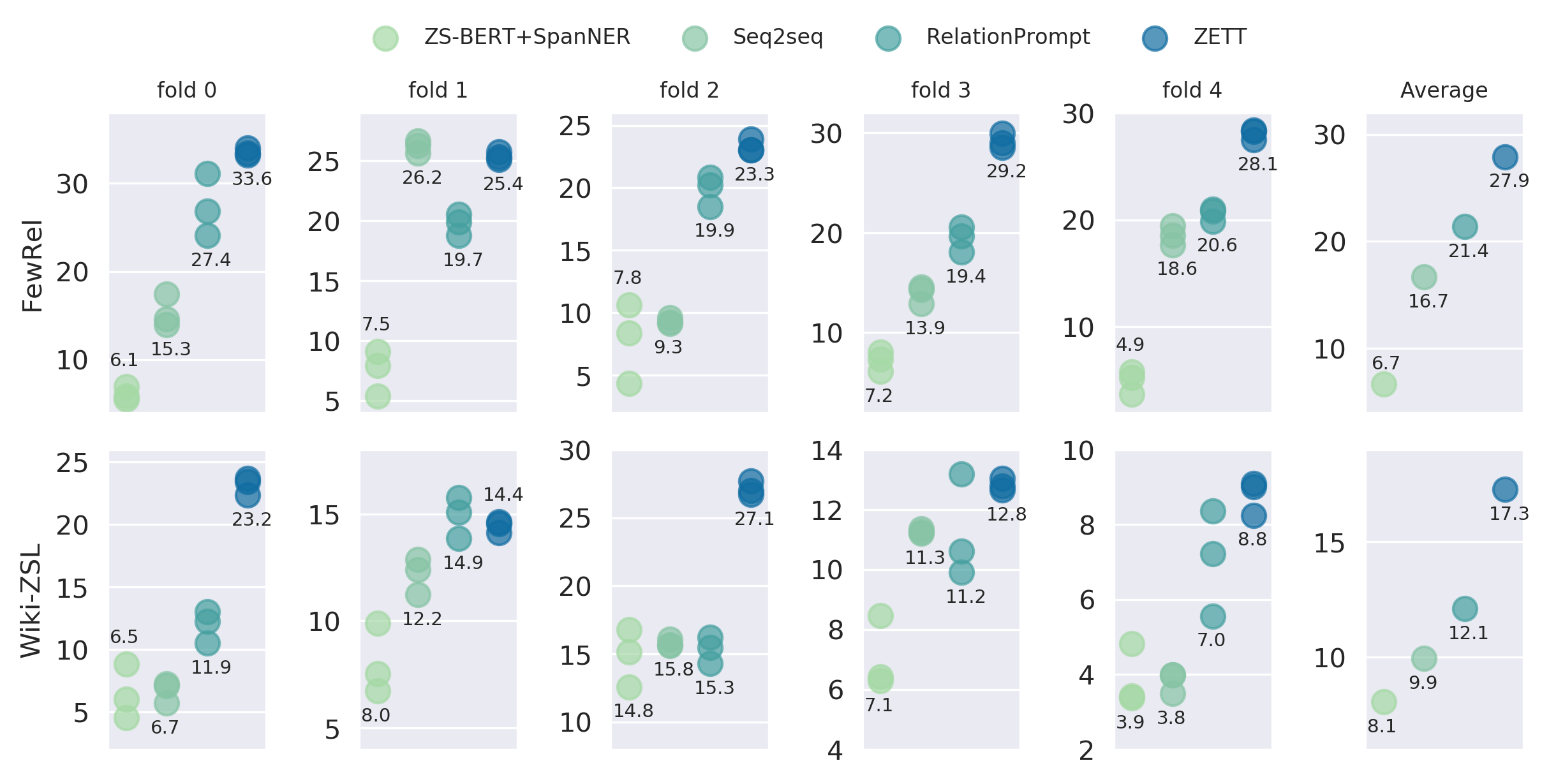}
    \caption{Results of each data fold in the single-triplet setting with $m$=10. We experiment on three different random seeds. Each point is the accuracy (\%) of each evaluation and labeled numbers are an average of three results.  Final column shows average over all folds.}
    \label{fig:perf_graph}
\end{figure*}

\begin{table*}[ht]
\centering
\small
\resizebox{\textwidth}{!}{
\begin{tabular}{c|l|ccc|ccc}
\toprule
\multicolumn{1}{c}{} & \multicolumn{1}{c}{} & \multicolumn{3}{c}{Single-triplet (Accuracy (\%))} & \multicolumn{3}{c}{Multi-triplet (F1 score (\%))}  \\
\multicolumn{1}{c}{} & \multicolumn{1}{c}{Model} & $m$=5 & $m$=10 & \multicolumn{1}{c}{$m$=15} & \multicolumn{1}{c}{$m$=5} & $m$=10 & $m$=15 \\ \midrule
\parbox[t]{2mm}{\multirow{5}{*}{\rotatebox[origin=c]{90}{FewRel}}} 
& TableSequence$^\dagger$ & 11.82 & 12.54 & 11.65 & 3.40 & 6.37 & 3.48 \\
& ZS-BERT+spanNER & 7.22 (±0.67) & 6.68 (±1.48) & 7.68 (±0.23) & 15.10	(±2.52)	&	12.14	(±2.33)	&	14.78	(±1.81) \\
& Seq2Seq & 22.02 (±1.07) & 16.71 (±0.82) & 11.91 (±0.70) & 26.02 (±2.81) & 17.12 (±1.86) & 13.02 (±0.96) \\
& RelationPrompt & 24.36 (±0.90) & 21.45 (±1.50) & 20.24 (±0.72) & 30.85 (±3.23) & 24.45 (±1.76) & 23.65 (±1.36) \\
& \textsc{ZETT} & \textbf{30.71} (±0.45) & \textbf{27.90} (±0.31) & \textbf{26.17} (±0.20) & \textbf{33.71} (±0.42) & \textbf{31.28} (±0.54) & \textbf{24.39} (±0.37) \\ \midrule
\parbox[t]{2mm}{\multirow{5}{*}{\rotatebox[origin=c]{90}{Wiki-ZSL}}}
& TableSequence$^\dagger$ & 14.47 & 9.61 & 9.20 & 6.29 & 6.4 & 6.39 \\
& ZS-BERT+spanNER & 8.16 (±1.26) & 8.05 (±0.79) & 6.47 (±0.42) & 15.96 (±1.11) & 11.98 (±2.28) & 11.90 (±0.84) \\
& Seq2Seq & 14.73 (±1.30) & 9.94 (±0.46) & 7.05 (±0.44) & 30.71 (±4.31) & 19.70 (±1.90) & 11.52 (±3.32) \\
& RelationPrompt & 16.74 (±1.53) & 12.13 (±0.86) & 10.47 (±0.96) & \textbf{33.28} (±1.70) & 24.04 (±2.12) & 18.73 (±1.98)  \\
& \textsc{ZETT} & \textbf{21.49} (±0.44) & \textbf{17.27} (±0.31) & \textbf{12.78} (±0.42) & 31.17 (±0.69) & \textbf{24.87} (±0.32) & \textbf{21.21} (±0.35) \\ \bottomrule
\end{tabular}}
\caption{Accuracy in the single-triplet and F1 score in the multi-triplet settings. $m$ denotes the number of unseen relations. All reported results are averaged over three different random seeds, where the result of each seed are averaged over five different data folds. Results of $\dagger$ are taken from \citet{chia-etal-2022-relationprompt}.}
\label{tab:main_results}
\end{table*}

\subsection{Experimental Settings}
\paragraph{Training}
We use the pre-trained T5-base\footnote{\url{https://huggingface.co/t5-base}}.
We fine-tune the model for 3 epochs with 64  batch size, and 3e-5 learning rate. We tune the parameters using the validation set. We provide details of the parameters in Appendix.

\paragraph{Inference}

At inference, ZETT generates entity spans given the input sentence concatenated with the relation template. We restrict the vocabulary to use tokens from the input sentence to ensure entities are spans from the input sentence.
We use a beam size of 4 to generate a maximum of 4 candidate entity pairs for each relation type. 
For the relation constraint, we set a threshold $\delta$ and choose $\delta$=0.85 based on the validation performance. 

\paragraph{Single- and Multi-triplet evaluation}
Each example in the datasets includes one or more triplets. We evaluate the models separately on single- and multi-triplet settings following the previous study~\cite{chia-etal-2022-relationprompt}. For the single-triplet setting, the examples have only one correct (gold) triplet, thus we use accuracy as the metric for evaluating performance.
In the multi-triplet setting, the number of gold triplets is two or more, thus we evaluate performance with a F1 score.
To retrieve positives in the multi-triplet setting, we set a threshold and output a candidate as a predicted positive example if its score is above this threshold. The thresholds for the multi-triplet evaluation are provided in Appendix.

\subsection{Baseline methods}
We compare the performance of ZETT with four existing methods for triplet extraction:
\textbf{1) ZS-BERT+spanNER} is a pipeline model, where two sub-tasks: relation classification and entity extraction are combined to extract triplets. We use ZS-BERT \citep{chen-li-2021-zs} for the relation classification model and implement span-based named entity recognition model using transformer encoder initialized with \texttt{roberta-base} checkpoint for entity extraction~\citep{fu-etal-2021-spanner}.
\textbf{2) TableSequence} \citep{wang-lu-2020-two} is a joint learning model with two separate encoders performing relation extraction and named entity recognition at the same time.
Since TableSequence is designed for supervised learning, we report the results of models trained on synthetic data from \citet{chia-etal-2022-relationprompt}.
\textbf{3) Seq2seq} \citep{chia-etal-2022-relationprompt} is an encoder-decoder  model based on the pre-trained BART-base \citep{lewis-etal-2020-bart}. The input to the encoder is a context, then the decoder generates a triplet as a sequence of structured template.
\textbf{4) RelationPrompt} \citep{chia-etal-2022-relationprompt} is an additionally fine-tuned model of the Seq2Seq on the synthetic data for the unseen relations.
\footnote{Since the synthetic data and pre-trained checkpoints for all experiments are not provided in the official implementation, we couldn't reproduce the results of RelationPrompt as same as reported in the paper. Thus, we re-trained the models with three different random seeds and report the average of them in Table \ref{tab:main_results}.}
They generate synthetic training datasets for unseen relations using GPT-2~\cite{radford2019language}\footnote{\url{https://huggingface.co/gpt2}} and fine-tune the Seq2seq-based triplet extraction model.

\begin{table*}[ht]
\centering
\small
\begin{tabular}{l|cc|cc}
\toprule
\multicolumn{1}{c}{} & \multicolumn{2}{c}{Single-triplet (Accuracy ($\Delta$))} & \multicolumn{2}{c}{Multi-triplet (F1 score ($\Delta$))}  \\
\multicolumn{1}{c}{Model} & FewRel & \multicolumn{1}{c}{Wiki-ZSL} & \multicolumn{1}{c}{FewRel} & Wiki-ZSL \\ \midrule
RelationPrompt & 21.45 & 14.37 & 24.45 & 24.04 \\
\text{ZETT} w/ \textit{Manual} 
& 27.90 & 17.27 & 31.28 & 24.87 \\
\text{ZETT} w/ \textit{Paraphrased} (Random) & 24.74 (-3.16) & 14.49 (-2.78) & 28.86 (-2.42) & 23.39 (-1.48) \\
\text{ZETT} w/ \textit{Paraphrased} (Top 1) & 25.12 (-2.78) & 15.34 (-1.93) & 29.60 (-1.68) & 24.63 (-0.24) \\\bottomrule
\end{tabular}
\caption{Comparison of performance with paraphrased templates at $m$=10. $\Delta$ is the performance difference over \text{ZETT} with manual templates.}
\label{tab:paraph_template}
\end{table*}

\section{Results}
\subsection{Automatic Evaluation}
\label{sec:main_results}
Table \ref{tab:main_results} shows the results of single- and multi-triplet evaluation settings on FewRel and Wiki-ZSL. As can be seen, ZETT consistently outperforms existing methods across different datasets under single-triplet setting. It achieves up to 6.45 and 5.14 higher accuracy than the existing state-of-the-art model, RelationPrompt on FewRel and Wiki-ZSL datasets, respectively. It also shows much lower variance in performance than RelationPrompt (see Figure \ref{fig:perf_graph}). In  the worst case, performance of  RelationPrompt differs by 7.1 points in accuracy (fold 0, $m$=10, FewRel). We conjecture that the variance can be attributed to the varying quality of the synthesized dataset in every trial. On the the other hand, ZETT shows stable performance through all trials, consistently outperforming the existing methods. In the multi-triplet setting, ZETT achieves the best F1 score for different relation set sizes except on Wiki-ZSL with $m$=5. We argue that this is mainly due to the biased distribution in the multi-triplet test sets; most examples are only for few relations. Therefore, performance loss in a particular relation results in significant drop in overall performance. We analyze the main causes of prediction failure in Section \ref{sec:error_analysis}. Overall, results of automatic evaluation show that having a simpler training process can yield more effective and stable performance on the task. 

\subsection{Human Evaluation}
\label{sec:human_eval}
{Both the test sets from our experiments are created using distant supervision and hence can be noisy and incomplete. In other words, they can include triplets that are not supported by an input text. Furthermore, it may not include all possible triplets from an input text. We, therefore, conduct human evaluation to better understand the performance of ZETT. We focus on WikiZSL since it is noisier and sample 200 contexts for manual annotation. We ask three CS graduates to annotate top-5 predictions for each context. An annotator labels a prediction correct if it is supported by the input text.
Three annotators labeled the triplets such that each triplet receives two annotations. 
We identified triplets labeled as \textit{True} by both annotators, showing a high agreement with a Cohen's Kappa coefficient of 0.75. Among the 1,000 triplets, we found 127 mislabeled instances--24 were \textit{False} (i.e., unsupported by input text) but originally labeled as \textit{True}, and 103 were \textit{True} but not covered in the original dataset.
We found that accuracy of ZETT increased from 18\% to 30.2\% on this manually annotated dataset. We will release the manually annotated dataset for future research.}

\section{Discussion} %
\label{sec:analysis}

\subsection{Robustness to choice of templates}
Since ZETT uses templates to generalize to unseen relations, the performance can be sensitive to the
wording %
of templates~\cite{sanh2022multitask}. In this section, we investigate how the performance varies depending on the template.
{To that end, we paraphrase the manually defined templates and compare the performance of ZETT with manual and paraphrased templates.}

We adopt the back-translation method from \citet{jiang-etal-2020-know}, which uses an English-German machine translation model to generate semantically-similar templates. {In our experiments, we used the translation model to generate 7 templates in German language for each manual template, and then back-translate each of them to 7 templates in English. As a result, we obtain 49 paraphrased templates per relation. Since many of them are duplicates, we evaluate on the most-frequent template, \textit{Paraphrased} (Top 1). We also compare with a randomly selected paraphrased template, \textit{Paraphrased} (Random). We will release the full set of paraphrased templates.}

\begin{table*}[t]
\centering
\resizebox{\textwidth}{!}{
\begin{tabular}{l|l|l}
\toprule
\multicolumn{3}{l}{\textsc{Context}: \textcolor{blue}{Jimmy Jam} is the son of \textcolor{purple}{Cornbread Harris}, a Minneapolis blues and jazz musician.} \\
\multicolumn{3}{l}{\textsc{Gold Triplet}: (\textcolor{blue}{Jimmy Jam}, \textit{father}, \textcolor{purple}{Cornbread Harris})} \\ \midrule
\multicolumn{1}{l|}{\multirow{2}{*}{RP}} & \textsc{Model output} & Head Entity : Cornbread Harris , Tail Entity : Minneapolis blues, Relation : field of work . \\ \cline{2-3}
\multicolumn{1}{l|}{}                    & \textsc{Triplet} & (\textcolor{purple}{Cornbread Harris}, field of work, Minneapolis blues) \\ \midrule
\multirow{3}{*}{ZETT}  & \textsc{Template} & \texttt{<tail>} is a father of \texttt{<head>} \\ \cline{2-3}
                       & \textsc{Model output} & \texttt{<X>} Cornbread Harris \texttt{<Y>} Jimmy Jam \texttt{<Y>} \\ \cline{2-3}
                       & \textsc{Triplet} & (\textcolor{blue}{Jimmy Jam}, \textit{father}, \textcolor{purple}{Cornbread Harris}) \\ \midrule       
\multicolumn{3}{l}{} \\ \midrule
\multicolumn{3}{l}{\textsc{Context}: He participated in \textcolor{blue}{UEFA Euro 1972} for the \textcolor{purple}{Hungary national football team}.} \\
\multicolumn{3}{l}{\textsc{Gold Triplet}: (\textcolor{blue}{UEFA Euro 1972}, \textit{participating team}, \textcolor{purple}{Hungary national football team})} \\ \midrule
\multicolumn{1}{l|}{\multirow{2}{*}{RP}} & \textsc{Model output} & Head Entity : Hungary national football team , Tail Entity : UEFA Euro 1972, Relation : participating team . \\ \cline{2-3}
\multicolumn{1}{l|}{}                    & \textsc{Triplet} & (\textcolor{purple}{Hungary national football team}, \textit{participating team}, \textcolor{blue}{UEFA Euro 1972}) \\ \midrule
\multirow{3}{*}{ZETT}  & \textsc{Template} & \texttt{<tail>} is a participating team in \texttt{<head>} \\ \cline{2-3}
                       & \textsc{Model output} & \texttt{<X>} Hungary national football team \texttt{<Y>} UEFA Euro 1972 \texttt{<Y>} \\ \cline{2-3}
                       & \textsc{Triplet} & (\textcolor{blue}{UEFA Euro 1972}, \textit{participating team}, \textcolor{purple}{Hungary national football team}) \\ \bottomrule  
\end{tabular}}
\caption{Example \textsc{Model output} sequences and triplets from RelationPrompt (RP) and ZETT.}
\label{tab:analysis_eg}
\end{table*}

Table \ref{tab:paraph_template} compares the performance of ZETT with manual and paraphrased templates. {Since automatically paraphrased templates can be noisy in capturing the semantic meaning of a relation, we observe small performance drops when using ZETT with paraphrased templates. The performance drop is much smaller with \textit{Paraphrased} (Top 1). %
Even though the paraphrased templates are noisy, ZETT still consistently outperforms current state-of-the-art methods.}

\subsection{Ablation Study}
\label{sec:ablation_study}
Next, we conduct an ablation study to examine the importance of each generation setting and the relation constraint. The results are summarized in Table \ref{tab:ablation}. First, {we test without the vocabulary constraints that limits the vocabulary set to tokens that appear in the context. We observe a small drop in accuracy of up to 0.39 points. Second, we compare the performance of beam search with greedy decoding that only generates one sequence. We find that beam search improves accuracy by up to 1.62 points since it selects the best complete sequence of entity pair as opposed to selecting the best individual entity token in each position. Last, we observe that our proposed relation constraint, although simple, is effective in eliminating irrelevant relations and can improve accuracy by up to 3.56 points.}

\subsection{Qualitative Analysis}

{To further gain insights into the strengths and weaknesses of the models, we manually inspect examples where ZETT and RelationPrompt generate different triplets. Table \ref{tab:analysis_eg} shows some such examples. We observe that RelationPrompt often generates an incorrect triplet even for easy examples. For example, it fails to predict the correct relation \textit{father} even when a context (``Jimmy Jam is the son of Cornbread Harris'') clearly describes it. This can be attributed to noise in the synthesized training data for unseen relations. We find the training data for relation \textit{father} includes noisy examples such as ``The Last Days of the Lion-Cat is based on David Simon.'' $\rightarrow$ (The Last Days of the Lion-Cat, \textit{father}, David Simon), where the context has no information about \textit{father}. Such noisy examples can propagate errors in the multi-step training process of RelationPrompt. ZETT, being a single-step process, is more robust to such errors.}

{We also find that RelationPrompt often fails to predict the correct order of entities. This is because the output sequence for generating a triplet "Head Entity: \texttt{<head>}, Tail Entity : \texttt{<tail>}, Relation : \texttt{<relation>}" does not encode any information about the order of entities. In contrast, ZETT can leverage the implicit information about entity types and their order encoded in relation templates. For instance, for the relation \textit{participating team}, the template ``\texttt{<tail>} is a participating team in \texttt{<head>}'' provides implicit information that \texttt{<head>} entity is a sports team and \texttt{<tail>} entity should be a contest or sports game. This information helps ZETT correctly predict the order of entities.}

\subsection{Error Analysis}
\label{sec:error_analysis}

To understand the limitations of ZETT, we perform a detailed analysis of errors on 200 examples. Incorrect ranking of relations contributed to the most frequent errors, accounting for 36\% of errors. Since ZETT relies on LM's probability $P_{T5}$, it tends to assign high scores to relations that are frequently used in common sentences. We find that relations such as \textit{occupation}, \textit{owned by} or \textit{work location} had higher scores than more rare relations such as \textit{contains administrative territorial entity} or \textit{place served by transport hub}.
Lack of discriminatory power over semantically similar relations contributed to 21\% of the errors. For example, relations such as \textit{headquarters location}, \textit{location of formation}, and \textit{work location}\, all represent the concept of location and are hard for the model to discriminate without any fine-tuning.
Lastly, we find that relation constraint sometimes excluded the correct relation, contributing to 17\% of the errors. For instance, when the context ``Ay was the penultimate Pharaoh of Ancient Egypt's 18th dynasty.'' and the gold triplet (Ay, \textit{country of citizenship}, Ancient Egypt) are given, even though the context contains the information of Ay's citizenship, our model predict as (Ay,	\textit{occupation}, Pharaoh) by determining \textit{country of citizenship} is irrelevant.
The rest of error types include failure of predicting entity spans, flipped entity pair, generation of null string, and labeling errors of datasets. We provide more detailed examples of error types in Appendix. Future work can focus on developing more effective score functions and relation classification techniques.

\begin{table}[t]
    \centering
    \small
    \begin{tabular}{lcc}
    \toprule
         & FewRel & Wiki-ZSL \\\midrule
         ZETT & 27.90 & 17.27 \\
\quad w/o decoding vocab constraint & 27.76 & 16.88 \\
\quad w/o beam search & 26.87 & 15.65 \\
\quad w/o relation constraint & 24.34 & 16.61 \\ \bottomrule
    \end{tabular}
    \caption{Ablation study on $m$=10 and single-triplet setting. The metric is accuracy.}
    \label{tab:ablation}
\end{table}

\section{Related Work}

\paragraph{Zero- and few-shot Triplet Extraction}

Triplet extraction has largely been studied as a pipeline of two sub-tasks: entity extraction and relation extraction, with pipeline~\cite{zhong-chen-2021-frustratingly}, joint-learning \citep{roth-yih-2004-linear,yu-lam-2010-jointly,10.1145/2509558.2509559,miwa-sasaki-2014-modeling,li-ji-2014-incremental,tanl} and end-to-end neural approaches \citep{zheng-etal-2017-joint}.
Although open information extraction \citep{etzioni2008open} shares a similar objective, it extracts relation spans from the input text which later have to be canonicalized to obtain relation types. Triplet extraction instead targets a predefined set of relation types. Zero-shot triplet extraction aims to generalize the models to an unseen set of canonicalized relation types.
The task was proposed by \citet{chia-etal-2022-relationprompt} that generalizes by learning to create synthetic data for unseen relations. However, their approach does not provide guarantees of quality and consistency of the synthetic data, making it hard to reproduce.

Our task formulation that relies on generative models is inspired by recent progress in relation extraction. \citet{yang-etal-2021-entity} shows that leveraging additional information about entities can yield better zero-shot and few-shot performance on relation extraction. \citet{wang-etal-2021-zero} shows that a unified framework based on text-to-triple model can achieve good zero-shot performance for open information extraction and relation classification tasks.
Inspired by these observations, we propose a text-to-text approach for the triplet extraction task that leverages additional information encoded in relational templates to achieve state-of-the-art performance in zero-shot settings.

\paragraph{LM prompt-tuning with templates}
Recent progress in LM prompt-tuning aims to bridge the gap between pre-training and downstream tasks by using natural language templates. 
Most approaches reframe the downstream task as a masked language modeling problem and have been successfully applied for 
text classification \citep{obamuyide-vlachos-2018-zero,hu-etal-2022-knowledgeable,chen-etal-2022-meta}, named entity recognition \citep{cui-etal-2021-template}, and natural language inference \citep{schick-schutze-2021-exploiting,schick-schutze-2021-just}. 
However, these approaches are mostly tailored for classification tasks, and not suitable for structured prediction such as triplet extraction.
\citet{hsu-etal-2022-degree} introduced DEGREE, a method akin to ZETT, employing a template infilling task to extract events in a low-resource setting. However, DEGREE uses placeholders like ``someone'' or ``somewhere'' that are not part of the pre-training process, necessitating fine-tuning the model to new types of event. In contrast, the fine-tuning of ZETT is identical to pre-training of T5, that improves its capability to generalizability to unseen relations.

\section{Conclusion}
We introduced the ZETT, a new framework for zero-shot triplet extraction which does not need any data augmentation or pipeline systems. We reformulate the triplet extraction as a template infilling problem using natural language templates. This enables the model to better leverage PLMs by aligning pre-training, fine-tuning, and inference objectives and eliminates the need for additional training data for unseen relations. ZETT is effective in extracting triplet by leveraging knowledge in PLMs, and is also more stable with a simple training process. Through experiments on two datasets, we demonstrated that ZETT outperforms previous state-of-the-art methods, showing consistent performance improvement without any extra models or synthetic data. We also showed that ZETT is robust in the variation of templates, showing competitive results without a significant performance loss.
\section*{Limitations}
\paragraph{Method limitations}
As discussed in Section \ref{sec:error_analysis}, ZETT has several weaknesses:  1) the ranking score based on the PLM’s probability is likely high when the templates of the relation include phrases or sentences which are commonly used in corpus. 2) The model struggles to discriminate between similar relations. 3) The relation constraint method proposed in Section \ref{sec:relation_constraint} could exclude the relevant relations in inference.
Future work could explore more effective relation classification methods for the relation constraint and sophisticated score functions which are well generalized to relations even whose templates are infrequent in corpus.

\paragraph{Zero-shot setup limitations}
The zero-shot setup that has been explored in the literature assumes that the set of unseen relations is given in inference, and the number of unseen relations is at most 15. This setup cannot fully reflect the real-world problems where there are numerous unseen relations. Future work could include exploring more realistic task setups and developing techniques to overcome the challenges posed by them.

\bibliography{main}
\bibliographystyle{acl_natbib}

\clearpage
\appendix

\section{Appendix}

\subsection{Experimental Setup Details}

\begin{table}[ht]
    \centering
    \small
    \begin{tabular}{c|c}
    \hline
        batch size & 64 \\
        learning rate & 3e-5 \\
        warm-up ratio & 0.2 \\ 
        maximum input length & 128 \{128, 256\} \\
        maximum output length & 64 \{64, 128, 256\} \\
        beam size & 4 \{1, 2, 4, 8, 16\} \\
        \hline
    \end{tabular}
    \caption{Best-performing hyperparameters and search space. Values in parentheses denote search space.}
    \label{tab:hyps}
\end{table}

\begin{table}[ht]
    \centering
    \small
    \begin{tabular}{c|c|c}
    \hline %
        \multicolumn{3}{c}{Threshold} \\
        \multicolumn{1}{c}{$m$} & \multicolumn{1}{c}{FewRel} & \multicolumn{1}{c}{WikiZSL} \\ \hline
        5 & -2.6 & -2.6 \\
        10 & -2.5 & -2.5 \\
        15 & -2.5 & -2.6 \\
        \hline
    \end{tabular}
    \caption{Best-performing thresholds in the multi-triplet evaluation. The triplets with scores ($\log P_{T5}$) above the threshold are considered final predictions.}
    \label{tab:th-multi-triplets}
\end{table}

\begin{table}[ht]
    \centering
    \small
    \begin{tabular}{c|c}
    \hline
        Model & \# params \\ \hline
        ZS-BERT+spanNER & 224M \\
        TableSequence \citep{wang-lu-2020-two} & 240M \\ 
        Seq2seq \citep{chia-etal-2022-relationprompt} & 140M \\
        RelationPrompt \citep{chia-etal-2022-relationprompt} & 140M \\
        ZETT & 220M \\
        \hline
    \end{tabular}
    \caption{The number of parameters in each model.}
    \label{tab:model_size}
\end{table}

\paragraph{Hyperparameter}
Table \ref{tab:hyps} describes hyperparameters and search spaces we considered in experiments.
In training, we used AdamW optimizer \cite{loshchilov2018decoupled} for all transformer-based models with 0.2 warm-up ratio.

\paragraph{Thresholds of the multi-triplet evaluation}
As mentioned in Section \ref{sec:relation_constraint}, we use a threshold to retrieve positives when the example includes two or more triplets. We repeated the evaluation with the threshold in range of \{2.0, 2.1, 2.2, …, 3.4, 3.5\} on the validation set and chose the best performing ones. We report the detailed values in Table \ref{tab:th-multi-triplets}.

\paragraph{Computing Infrastructure}
We ran all experiments on a single NVIDIA GeForce GTX 1080 (12GB) with CUDA 10.1 version.

\paragraph{Computational Budget}
Training ZETT with hyperparameters in Table \ref{tab:hyps} takes 1.5h on the FewRel dataset and 2h on the Wiki-ZSL dataset with a single NVIDIA GeForce GTX 1080.

\paragraph{Model Parameters}
Table \ref{tab:model_size} provides the number of parameters in each model used in our experiments.

\subsection{Guidelines for Human Evaluation}
The goal of the human evaluation is to address that the datasets have some wrong labeled examples and false negatives. We manually evaluated 1,000 examples: Top 5 predictions (triplets) of ZETT from 200 contexts. The contexts are randomly sampled from the single-triplet test set of $m$=10 in Wiki-ZSL. Three annotators are given 667, 667, and 666 examples, respectively. Each example consists of input sequence (context and template), gold triplet, predicted triplet, and the model's score $P_{T5}$. The instructions for annotation are as follows:
\begin{itemize}
    \item 1) Annotate \textbf{TRUE} if you think the given context and the model’s prediction (triplet) are matched. Otherwise annotate \textbf{FALSE}.
    \item 2) Annotate \textbf{FALSE} if we cannot infer the triplet from the given context, even if the triplet itself is true.
    Concretely, for the example of 
the context ``Elected to the comptrollers post in 1998 as a Republican , Keeton ran as an independent candidate for Texas governor against Republican incumbent James Richard Rick Perry in 2006.'' and the triplet
(James Richard Rick Perry, \textit{residence}, Texas),
We cannot be sure whether James Richard Rick Perry lives in Texas or not just given the context. Thus, we should label this as \textbf{FALSE}.
\end{itemize}

We also provided the relation descriptions to avoid confusion between similar relations. For example, the relation \textit{residence} and \textit{location} can be confusing to annotators since both relations can refer to places of something. However, when we refer the description, we can clarify that these two relations have different head entity types: person for \textit{residence}, and \{object, structure or event\} for \textit{location}.

\begin{itemize}
    \item \textit{residence} (P551): the place where the person is or has been, resident
    \item \textit{location} (P276): location of the object, structure or event. In the case of an administrative entity as containing item use P131. For statistical entities use P8138. In the case of a geographic entity use P706. Use P7153 for locations associated with the object.
\end{itemize}

\begin{table*}[t]
\centering
\resizebox{\textwidth}{!}{
\begin{tabular}{l|l}
\hline
\multicolumn{2}{l}{a ) \textsc{Context}: This lizard lives in the southwestern part of Africa, in \textcolor{purple}{Namibia} and \textcolor{blue}{South Africa}.} \\
\multicolumn{2}{l}{\textsc{Labeled triplet}: (\textcolor{blue}{South Africa}, \textit{shares border with}, \textcolor{purple}{Namibia})} \\ \hline 
\multicolumn{2}{l}{b) \textsc{Context}: He studied photography at the \textcolor{blue}{International Center of Photography} in \textcolor{purple}{New York City} in 1990, under the tutelage} \\
\multicolumn{2}{l}{ of Larry Clark and Nan Goldin.} \\
\multicolumn{2}{l}{\textsc{Labeled Triplet}: (\textcolor{blue}{International Center of Photography}, \textit{headquarters location}, \textcolor{purple}{New York City)}} \\ \hline
\multicolumn{2}{l}{c) \textsc{Context}: FC Slavutych was a \textcolor{blue}{Ukrainian} football club from Slavutych, \textcolor{purple}{Kiev Oblast}.} \\
\multicolumn{2}{l}{\textsc{Labeled Triplet}: (\textcolor{blue}{Ukrainian}, \textit{contains administrative territorial entity}, \textcolor{purple}{Kiev Oblast})} \\ \hline
\multicolumn{2}{l}{d) \textsc{Context}: Google released \textcolor{purple}{Android} 7.1.1 Nougat for the \textcolor{blue}{Pixel C} in December 2016.} \\
\multicolumn{2}{l}{\textsc{Labeled Triplet}: (\textcolor{blue}{Pixel C}, \textit{operating system}, \textcolor{purple}{Android})} \\ \hline
\end{tabular}}
\caption{Wrong labeled examples in the FewRel dataset. In example a), b), and c), although these triplets are true, the contexts are not related to the given relations or the triplets cannot be inferred from the contexts. Another type of error is a false negative. In the example d), our model predicts (Pixel C, \textit{owned by}, Google) given the context, and this knowledge is also true. However, the triplet (Pixel C, \textit{owned by}, Google) is not labeled as an answer in the dataset.}
\label{tab:eg_wrong_labeled}
\end{table*}

 \begin{figure*}[t]
    \centering
    \includegraphics[width=1.0\linewidth]{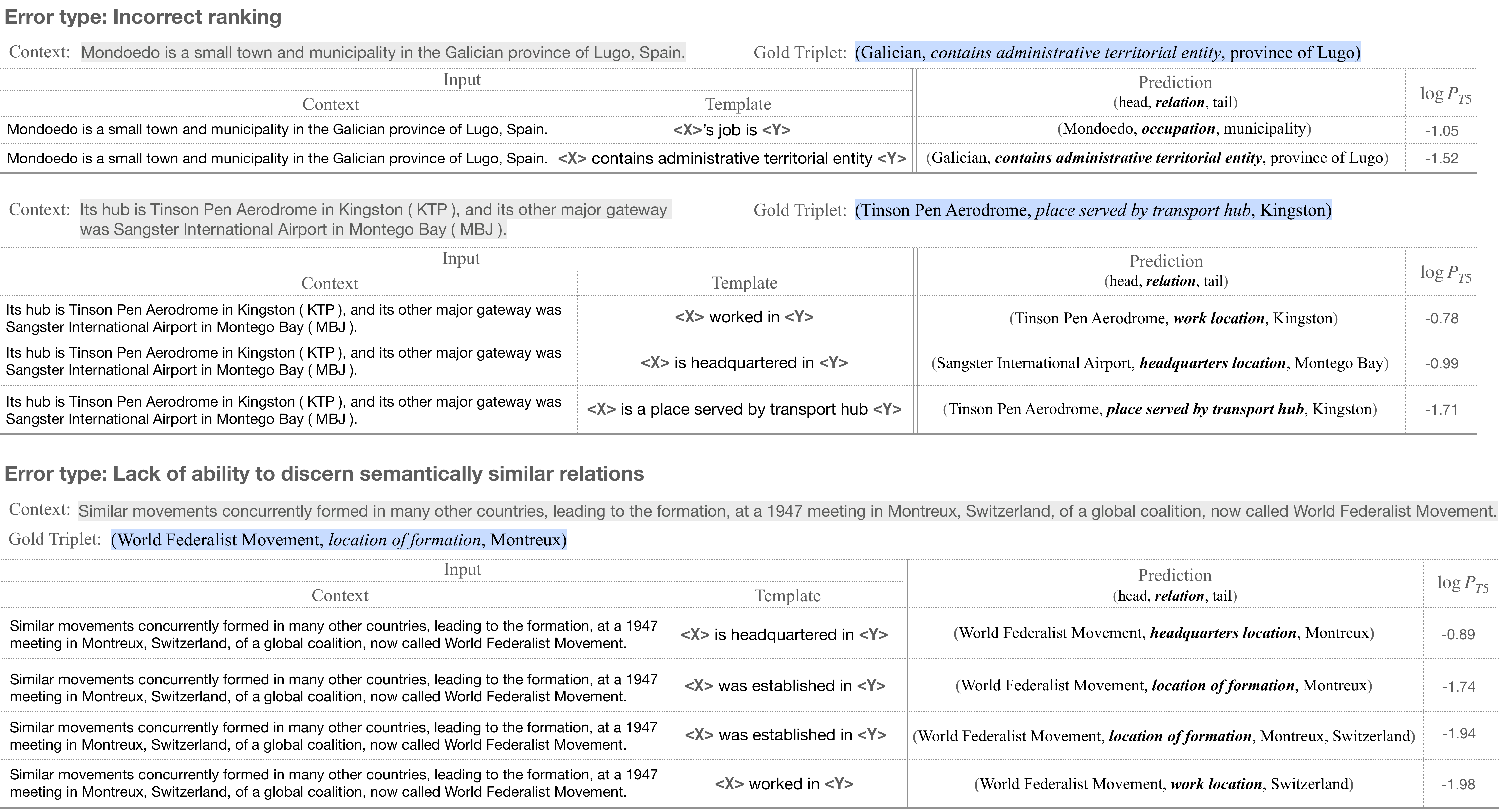}
    \caption{Examples according to error types. The first line of each table denotes the final prediction.}
    \label{fig:error_analysis}
\end{figure*}

\end{document}